\documentclass[sigconf,natbib=false]{acmart}

\usepackage{lipsum}
\usepackage[ruled,vlined,linesnumbered]{algorithm2e}

\usepackage{tikz}
\usepackage{pgfplots}
\acmConference[]{Recsys '25: CONSEQUENCES Workshop}{September, 2025}{Prague}
\acmISBN{}

\RequirePackage[
  datamodel=acmdatamodel,
  style=acmnumeric,
  ]{biblatex}

\usepackage{subcaption}
\begin{document}

% \title{Non-linear  Aggregate Optimization allows going beyond expectation}
\title{Non-Linear Counterfactual Aggregate Optimization}

\author{Benjamin Heymann}
\authornote{Both authors contributed equally to this research.}
\affiliation{%
  \institution{Criteo AI Lab}
  \city{Paris}
  \country{France}}
\email{b.heymann@criteo.com}

\author{Otmane Sakhi}
\affiliation{%
  \institution{Criteo AI Lab}
  \city{Paris}
  \country{France}}
\authornotemark[1]
\email{o.sakhi@criteo.com}

\renewcommand{\shortauthors}{Heymann \& Sakhi}

\begin{abstract}
We consider the problem of directly optimizing a non-linear function of an outcome, where this outcome itself is the sum of many small contributions. The non-linearity of the function means that the problem is not equivalent to the maximization of the expectation of the individual contribution.
By leveraging the concentration properties of the sum of individual outcomes, we derive  a scalable descent algorithm   that  directly optimizes for our stated objective. This allows for instance to maximize the probability of
successful A/B test, for which it can be wiser to target a success criterion—such as exceeding a given uplift—rather than chasing the highest expected payoff.
\end{abstract}

\begin{CCSXML}
<ccs2012>
   <concept>
       <concept_id>10010405.10010481.10010484.10011817</concept_id>
       <concept_desc>Applied computing~Multi-criterion optimization and decision-making</concept_desc>
       <concept_significance>500</concept_significance>
       </concept>
   <concept>
       <concept_id>10002950.10003648.10003670.10003684.10003685</concept_id>
       <concept_desc>Mathematics of computing~Bootstrapping</concept_desc>
       <concept_significance>500</concept_significance>
       </concept>
 </ccs2012>
\end{CCSXML}

\ccsdesc[500]{Applied computing~Multi-criterion optimization and decision-making}
\ccsdesc[500]{Mathematics of computing~Bootstrapping}

\keywords{offline policy optimization, bootstrapping, non-linear optimization}


\maketitle

\section{Introduction}

Offline contextual bandit \cite{dudik2011doubly} is a widely used framework that leverages logged data from past interactions to improve future decision-making \cite{bottou2013counterfactual}. In classical off-policy optimization, the performance of any new policy $\pi$ is measured by its value $V(\pi)$, which is the expected payoff or reward obtained by playing actions according to $\pi$. Motivated by real world decision making problems, we look beyond the expected reward. We consider the problem of maximizing over the contextual policy $\pi$ the expectation of a general criterion $j$
\begin{align}
\label{eq:criteria}
    \mathbb{E}_{N, \{X_i \sim \nu, A_i \sim \pi(\cdot|X_i)\}_{i \in [N]}}\Bigg[j\Big(\sum_{i=1}^N R(X_i,A_i)\Big)\Bigg],
\end{align}
where $j:\mathbb{R}\to\mathbb{R}$ is a monotone, possibly discontinuous, function  over the aggregated outcome $\sum_{i=1}^N R(X_i,A_i)$. Each $x_i$ is a context coming from an unknown distribution $\nu$, --- in the setting of  Recommender Systems for instance, it would be the information known about the user ---  $N$ is a random integer that represents the number of individual experiments, $A_i$ are the actions played by $\pi$ for context $X_i$ and $R$ is a random, positive rewards. To perform this task~\eqref{eq:criteria}, the Decision Maker (DM) disposes of an offline dataset $(X_i,A_i,R_i)_{i\in 1\ldots N_0}$ generated by a policy $\pi_0$.
Otherwise said:  a policy is applied to a large population and only the distribution of the aggregated result matters.
 This formulation is notably generic. For instance, it can 
 handle hard constraints and  account for variance and risk aversion.
It includes the cases of many industrial applications, in particular RecSys. 
Typically, the DM is interested in the policy performance overall, and might want to trade-off robustness and performance at this aggregated level. Our work is motivated by the observation that instead of optimizing directly for the true objective, most methods rely on proxy goals such as maximizing the expected reward under some pessimism constraints or penalty. This framework is general and recovers \emph{expected value optimization} for instance when the criterion $j$ is set to the identity function.

\paragraph{\textbf{Costly A/B testing}} This is one of the motivating examples of this work. The DM is an engineering team who wants to maximize the probability of the designed policy to result in an A/B test to be positive according to some external criteria, say an uplift being above a given threshold.
The need to ensure the test is positive comes from the fact that A/B tests in large  systems are most of the time  resource demanding (i.e. Monitoring, A/B test slots, Risk). In such a scenario, the non-linear function $j$ could be threshold function
 \begin{align}
j(x) = \begin{cases}
    1 & \text{if } x \geq \bar{x} \\
   0 & \text{otherwise}.
\end{cases}
 \end{align}
 Here $\bar{x}$ is the bar to reach for a test to be deemed positive. Hence in this example, the objective~\eqref{eq:criteria} is to find a policy that maximizes the probability that the A/B test is positive.

\section{Setting}

We work under the offline contextual bandit framework \cite{bottou2013counterfactual}. The users' contexts $X_i$ are i.i.d. copies of a random variable coming from a fixed, unknown distribution $\nu$. These contexts are revealed to the system upon the user's arrival. The system is represented by a parameterized policy $\pi_\theta\,, \theta \in \Theta$, that given a context $X$, samples an action $A_\theta \sim \pi_\theta(\cdot|X)$, and then receives an outcome $Y_\theta$, modeled as a positive reward $Y_\theta = R(A_\theta, X) \in \mathbb{R}^+$. We are interested by the classical, off-policy learning setup, described as follows: 
\begin{enumerate}
    \item The DM receives a dataset $\mathcal{D}_n = \{X_i,A_i,R_i\}_{i\in [n]}$ collected by a policy $\pi_{0}$, where $n$ is random \footnote{In practical scenarios, $n$ is modeled as a Poisson.}.
    \item Leveraging $\mathcal{D}_n$, the DM learns a new policy $\pi_\theta$ to deploy.
    \item The variables $N$ and $(Y^1_\theta,Y^2_\theta,\ldots, Y^N_\theta)$ are then observed.
    \item The DM receives the payoff  $ j(H_{\theta})$, where 
    \begin{align*}
    H_\theta = H(Y^1_\theta,Y^2_\theta,\ldots, Y^N_\theta)=\sum_{i=1}^N Y^i_\theta,
\end{align*}
\end{enumerate}
so that objective~\eqref{eq:criteria} becomes
\begin{align*}
   \max_{\theta\in\Theta}\quad J(\theta) =  \mathbb{E}\left( j(H_{\theta})\right).
\end{align*}

% Let $(\Omega,\mathcal{F},\mathbb{P})$ be a probability space. 
% The users' contexts
% are i.i.d copies of a  random variable $X:\Omega \to \mathcal{X}$, revealed to the system upon the user's arrival. 
% For each context $X(\omega)$, the system
%  samples an action $A_{\theta}(X(\omega))$ from  a distribution $\pi_\theta(X(\omega))$, where $\pi_\theta$ belongs to a set of policies parametrized by $\theta\in \Theta$.
% The decision maker then observes an outcome $Y_{\theta} = F(A_{\theta}(X(\omega)),\omega)= R(X,A)$.

% The full setup is as follows: 
% \begin{enumerate}
%     \item The DM receives a sample $(X_i,A_i,R_i)_{i\in 1\ldots N_0}$ generated by a policy $\pi_{0}$,  where $N_0$  follows a Poisson distribution
%     \item The DM decides a policy $\pi_\theta$
%     \item The variables $N$ and $(Y^1_\theta,Y^2_\theta,\ldots, Y^N_\theta)$ realize
%     \item The DM receives the  payoff  $ j(H_{\theta})$, where 
%     \begin{align*}
%     H_\theta = H(Y^1_\theta,Y^2_\theta,\ldots, Y^N_\theta)=\sum_{i=1}^N Y^i_\theta,
% \end{align*}
% \end{enumerate}

\noindent The optimized objective is defined under actions coming from the new policy $\pi_\theta$. In practice, we want to learn in step (2) a policy $\pi_\theta$ that maximizes this objective only leveraging $\mathcal{D}_n$. The difficulty arises from the fact that the decision maker does not know the underlying distributions ($\nu, R, N$), $\mathcal{D}_n$ is collected under another policy $\pi_0$ and the criterion $j$ can be non-differentiable.

Here is how we plan to address those difficulties.
First, given $\mathcal{D}_n$, we can build an estimator of the aggregated payoff, using standard inverse propensity scoring \cite{horvitz1952generalization}:
$$H_\theta = \sum_{i = 1}^n \frac{\pi_\theta(A_i|X_i)}{\pi_0(A_i|X_i)}R_i.$$

If $\pi_\theta$ does not deviate extremely from $\pi_0$, $H_\theta$ enjoys a finite variance and it is reasonable to invoke the Central Limit Theorem (CLT)  \cite{clt}, and use the following approximation:
\begin{align*}
H(Y^1_{\theta},Y^2_{\theta},\ldots, Y^{N}_{\theta})\sim
\mathcal{N}(\mu_\theta,\sigma_\theta)
\end{align*}
where $\mathcal{N}$  refers to the Gaussian family, and $\mu_\theta$ and $\sigma^2_\theta$ the empirical mean and variance of $H_\theta$. We hence get the following approximation for  criterion~\eqref{eq:criteria}:
\begin{align}
\label{eq:newobjective}
    \hat{J}(\theta) =\mathbb{E}_{h\sim \mathcal{N}(\mu_\theta,\sigma_\theta) } [j\left(h\right)].
\end{align}

\noindent Observe that if $j$ induces risk aversion, then big importance weighting factors will be avoided, keeping the CLT approximation valid. 

\section{Related work}

\textbf{Inverse Propensity Scoring (IPS.)}
IPS is the go-to method for counterfactual estimation~\cite{horvitz1952generalization, bottou2013counterfactual}. It produces, under mild assumptions, an unbiased estimate of the average effect of a new policy $\pi$ using logged data generated by a given policy $\pi_0$. 
In the presence of linear aggregation, for example if $j$ is the identity, then:
\begin{align}
\label{eq:IPS}
   H(\pi_\theta) = \frac{1}{n} \sum_{i=1}^n R_i\frac{\pi_{\theta}(a^i|X^i)}{\pi_{0}(a^i|X^i)},
\end{align}
is an unbiased estimate of the expected reward. While IPS is an extremely powerful tool, it can suffer from large variance in practice \cite{metelli2021subgaussian}. Regularised IPS, often through clipping \cite{bottou2013counterfactual, swaminathan2015batch} or smoothing \cite{metelli2021subgaussian, aouali2023exponential, sakhi2024logarithmic} is used to trade bias for reduced variance. 

\textbf{Pessimism in off-policy learning.} Building on the idea that IPS is unreliable \cite{metelli2021subgaussian}, 
Recent approaches optimize empirical upper bounds on policy risk \cite{pmlr-v202-sakhi23a, sakhi2024logarithmic}. The idea is to evaluate a policy \emph{expected performance} under (high probability) worst-case conditions.

\textbf{Metapolicy.}
The approach developed in \cite{betlei2024maximizing} is closely related to our method. \cite{betlei2024maximizing} introduce an optimization problem to directly maximize the probability of success of a test by assigning buckets of user populations to policies. At the difference to our work, \cite{betlei2024maximizing} only consider the thresholding criteria, use a finite set of policies, and rely on a bucketization of the set of users.

\section{Gaussian approximation}

The CLT, and thus the gaussian approximation is the backbone of our approach. Once $n$ is large, and importance weights are controlled, the argument of $j$ in~\eqref{eq:newobjective} behaves like a Gaussian, allowing the estimation of the aggregated outcome variance from the data.

The Gaussian approximation allows us to gain one level of smoothness (without this approximation, the objective is not even continuous), and to use a gradient descent algorithm to optimize the policy $\pi_\theta$.
As a result, compared to the approach in~\cite{betlei2024maximizing}, our method does not require a bucketization of the users, and can be applied to a continuous policy class (as opposed to a finite set of policies).

One might wonder how this method addresses the potentially high variance associated with "crude" IPS.
While it is easy to design situation where this approach will dramatically fail (convex $j$ or extremely high threshold), we argue that there are many settings where the risk aversion induced by $j$ prevents the algorithm from using large importance weights, controlling the variance of the aggregated outcome.
Though concave function $j$ naturally induces this effect, other functions, such as certain threshold functions, can achieve similar outcomes under suitable conditions.

By applying standard chain rule arguments \cite{10.1007/BF00992696}, we obtain the descent algorithm described in Algorithm~\ref{alg:COGC}.

\RestyleAlgo{ruled}
\begin{algorithm}
\caption{\textbf{Counterfactual Aggregate Optimization}}\label{alg:COGC}
\textbf{Input}: Parameterized policy $\pi_\theta$, learning rate $\eta >0$, $m \ge 1$ number of gaussian samples. \\
\textbf{Initialize}: $\theta =\theta_0$. \\
\For{$k \geq 0$}{
Estimate $\mu_k = \mu_{\theta_k}$ and $\sigma^2_k = \sigma^2_{\theta_k}$ from the data. \\
Sample $n$ gaussian samples $h_1, \cdots, h_m \sim \mathcal{N}(\mu_{\theta_k}, \sigma^2_{\theta_k})$.\\ 

Compute a gradient estimate $\nabla_{\theta = \theta_k}\hat{J}(\theta)$: 
$$ \frac{1}{m \sigma^2_k} \sum_{\ell = 1}^m \left(\left(h_\ell - \mu_k \right) \nabla_\theta \mu_\theta + \frac{1}{2}\left((\frac{h_\ell-\mu_k}{\sigma_k} )^2-1 \right) \nabla_\theta \sigma^2_\theta \right)j(h_\ell) $$ \\

$\theta_{k+1} \leftarrow  \theta_k + \eta \nabla_{\theta = \theta_k}\hat{J}(\theta)$ .\\

}
\end{algorithm}
\vspace{-2mm}

\section{Experiments}
\begin{table*}
  \centering
  \begin{tabular}{ccccccc}
    \toprule
    Method & $\mathbb{E}[r]$  & $\mathbb{M}[r]$ & $\mathbb{P}(I > 10\%)$ & $\mathbb{P}(I > 20\%)$ & $\mathbb{P}(I > 30\%)$ & $\mathbb{P}(I < 0.)$ \\
    \midrule
    \texttt{IPS} & $0.062$ & $0.060$ & $0.69$ & $0.54$ & $0.47$ & $0.31$\\
    \texttt{LS} & $\boldsymbol{0.066}$ &  $0.066$ & $\boldsymbol{0.99}$ & $0.88$ & $\boldsymbol{0.59}$ & $\boldsymbol{0.}$\\
    \midrule
     $j(I > 10\%)$& $\boldsymbol{0.067}$ & $\boldsymbol{0.068}$ & $\boldsymbol{0.99}$ & $\boldsymbol{0.93}$ & $\boldsymbol{0.58}$ & $\boldsymbol{0.}$\\
     $j(I > 20\%)$& $\boldsymbol{0.067}$ & $\boldsymbol{0.068}$ & $\boldsymbol{0.99}$ & $\boldsymbol{0.93}$ & $\boldsymbol{0.58}$ & $\boldsymbol{0.}$\\
     $j(I > 30\%)$& $\boldsymbol{0.067}$ &  $\boldsymbol{0.068}$ & $\boldsymbol{0.99}$ & $0.91$ & $\boldsymbol{0.60}$ & $\boldsymbol{0.}$\\
    \bottomrule
  \end{tabular}
  \vspace{0.2cm}
  \caption{Performance of learned policies $\pi_\theta$. Optimizing the criteria is robust, LS is competitive and IPS is unreliable.}
  \label{tab:results}
  \vspace{-4mm}
\end{table*}

We validate the idea on the following synthetic example. 
The model is a single context, multi-armed bandit with $K = 1000$ actions and binary, bernoulli rewards $r \in \{0, 1\}$. We collect data using a skewed behavior policy $\pi_0$, putting more mass on actions of smaller indices than the others. This policy collects $N = 1000$ observations, producing a simple simulation, yet a hard instance where importance weighting approaches fail without proper variance control. 

In this setting, we investigate two classes of criteria $j$, $j(x)=x^\kappa$ for $0<\kappa<1$ and $j(x)=1\{x>\bar{x}\}$  for some threshold $\bar{x}$. For all methods, we use the class of softmax policies over possible actions:
\begin{align}
    \pi_\theta(A=i) \propto \exp({\theta_i}), \quad \theta\in\mathbb{R}^K
\end{align}
We use Algorithm~\ref{alg:COGC} to optimize our criteria, and we compare our method to two baselines: learning a policy using IPS \cite{horvitz1952generalization} and the \emph{pessimistic} Logarithmic Smoothing (LS) estimator \cite{sakhi2024logarithmic} with the smoothing parameter $\lambda$ set to its theoretical value.

\begin{figure}[ht]
  \centering

    \includegraphics[width=0.72\linewidth]{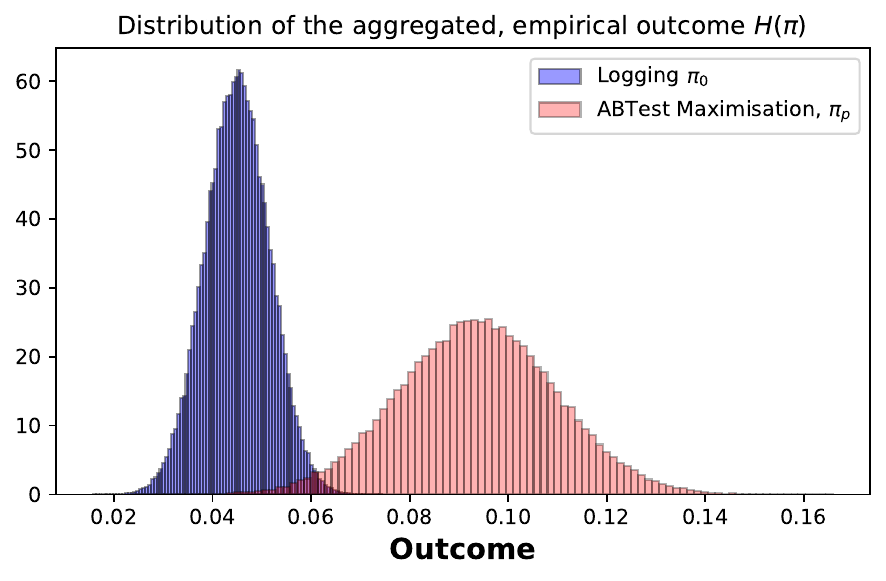}
  \vspace{-1mm}
  \caption{Empirical distribution of the learned policy through A/B test maximization.}
  \label{fig:abtest}
  \vspace{-1mm}
\end{figure}

\paragraph{\textbf{In-sample behavior}} To build intuition of our novel approach, we examine the empirical aggregated outcome $H_n(\pi_\theta)$ for the learned policy $\pi_\theta$ using $j(x) = 1[x > H_n(\pi_0)]$, which is the criteria maximizing the probability of improving on $\pi_0$. Figure~\ref{fig:abtest} plots the distribution of the outcome for the A/B test maximization policy $\pi_\theta$ compared to $\pi_0$. We can see that the new distribution of outcomes moved \emph{just enough} from the outcome distribution of $\pi_0$, maximizing the probability of improving $\pi_0$, which means increasing the reward, and controlling its variance so as to minimize the overlap of the two outcome distributions. 

\begin{figure}[ht]
  \centering
  \begin{subfigure}[t]{0.49\linewidth}
    \centering
    \includegraphics[width=\linewidth]{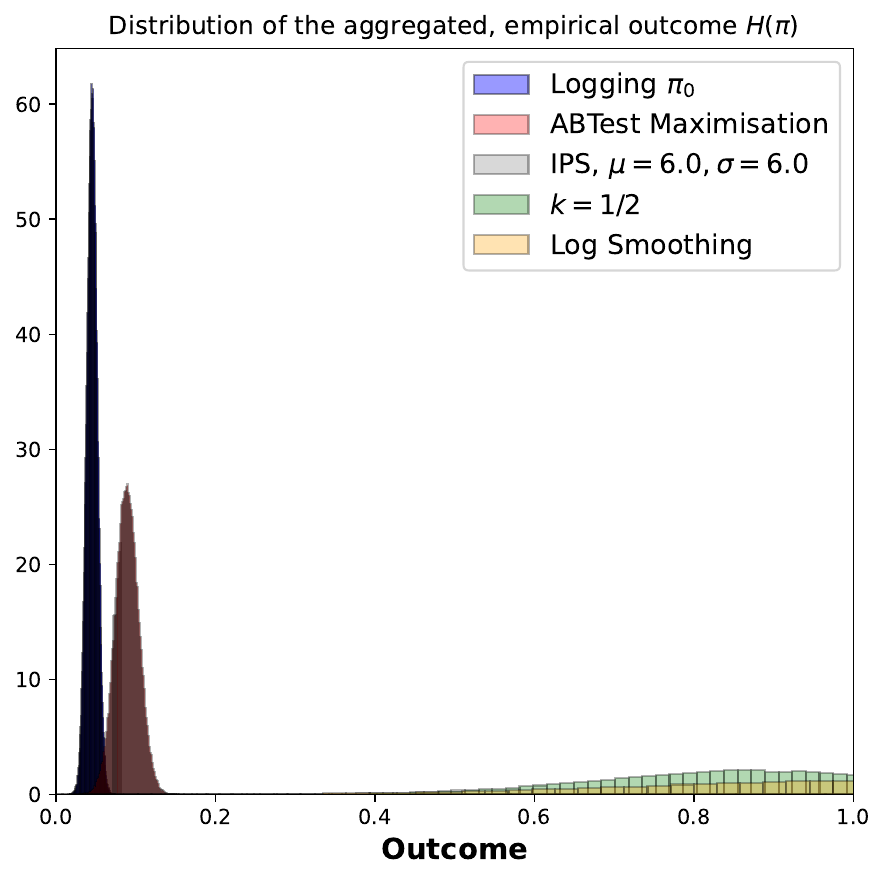}
  \end{subfigure}
  \hfill
  \begin{subfigure}[t]{0.49\linewidth}
    \centering
    \includegraphics[width=\linewidth]{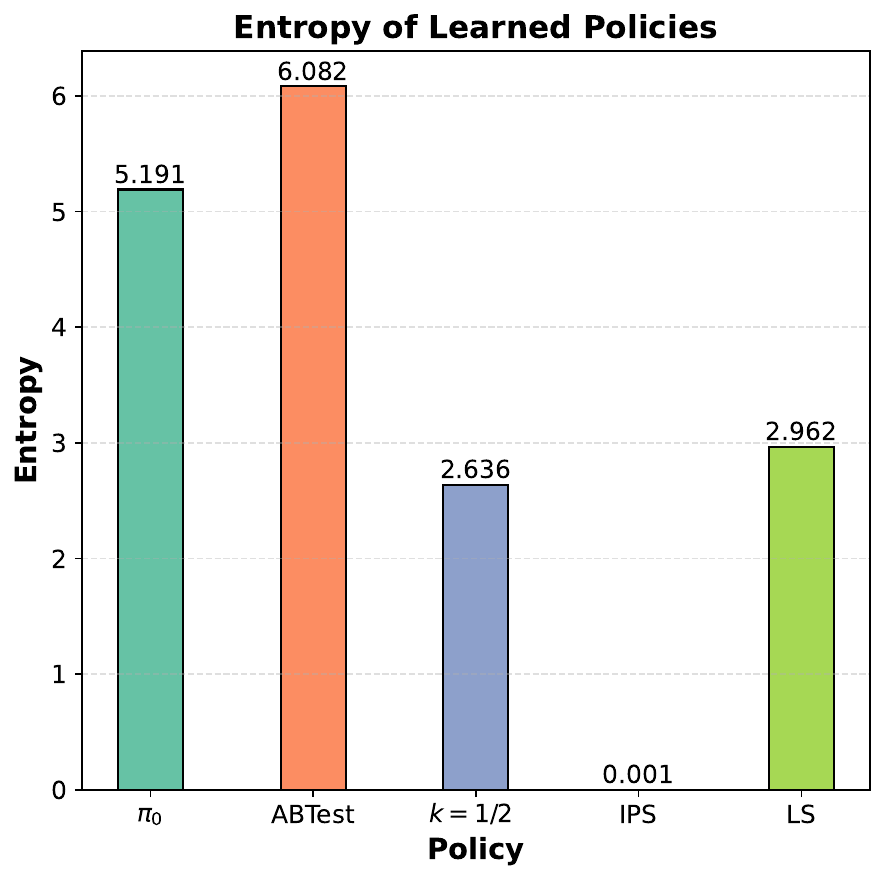}
  \end{subfigure}
  \vspace{-3mm}
  \caption{Comparative analysis of learned policies. Left: empirical outcome distribution $H_n(\pi)$. Right: entropy of learned policies $\pi_\theta$. Observe that IPS is over-confident.}
  \label{fig:combined}
    \vspace{-4mm}
\end{figure}

In Figure~\ref{fig:combined}, we additionally plot the distribution obtained for the policy $\pi_\theta$ maximizing criteria $j(x) = \sqrt{x}$, as well as the policies obtained by optimizing IPS and LS. Looking in the left plot, the first observation is that IPS is overly-confident, predicting an impossible outcome and suffers an incredibly large variance, making it unreliable in these conditions. Our criteria with $k = 1/2$ as well as the LS estimator induce similar risk aversion, still maximizing the aggregated outcome, obtaining distributions with high rewards but large variances. The right plot in Figure~\ref{fig:combined} displays the entropies of these policies. IPS converges to a nearly deterministic policy, consistently choosing the same action. However, our risk-averse criteria and LS both encourage a form of policy hedging. These methods converge to policies that play a diverse set of actions, which are still good enough to increase the reward of $\pi_0$.

% \begin{figure*}[t]
%   \centering
%   \begin{subfigure}[t]{0.49\linewidth}
%     \centering
%     \includegraphics[width=\linewidth]{graphs/all_methods_distribution.pdf}
%   \end{subfigure}
%   \hfill
%   \begin{subfigure}[t]{0.49\linewidth}
%     \centering
%     \includegraphics[width=\linewidth]{graphs/entropy_of_learned_policies.pdf}
%   \end{subfigure}
%   \vspace{-2mm}
%   \caption{Comparative analysis of learned policies. Left: empirical outcome distribution $H(\pi)$. Right: entropy of learned policies $\pi_\theta$. Observe that IPS is over-confident.}
%   \label{fig:combined}
% \end{figure*}

% \begin{table}
%   \label{tab:freq}
%   \begin{tabular}{ccccccc}
%     \toprule
%     Method & $\mathbb{E}[r]$  & $\mathbb{M}[r]$ & $\mathbb{P}(I > 0.1)$ & $\mathbb{P}(I > 0.2)$ & $\mathbb{P}(I > 0.3)$ & $\mathbb{P}(I < 0.)$ \\
%     \midrule
%     \texttt{IPS} & $0.062$ & $0.060$ & $0.69$ & $0.54$ & $0.47$ & $0.31$\\
%     \texttt{LS} & $\boldsymbol{0.066}$ &  $0.066$ & $\boldsymbol{0.99}$ & $0.88$ & $\boldsymbol{0.59}$ & $\boldsymbol{0.}$\\
%     \midrule
%      $j(I > 0.1)$& $\boldsymbol{0.067}$ & $\boldsymbol{0.068}$ & $\boldsymbol{0.99}$ & $\boldsymbol{0.93}$ & $\boldsymbol{0.58}$ & $\boldsymbol{0.}$\\
%      $j(I > 0.2)$& $\boldsymbol{0.067}$ & $\boldsymbol{0.068}$ & $\boldsymbol{0.99}$ & $\boldsymbol{0.93}$ & $\boldsymbol{0.58}$ & $\boldsymbol{0.}$\\
%      $j(I > 0.3)$& $\boldsymbol{0.067}$ &  $\boldsymbol{0.068}$ & $\boldsymbol{0.99}$ & $0.91$ & $\boldsymbol{0.60}$ & $\boldsymbol{0.}$\\
%   \bottomrule
% \end{tabular}
% \vspace{0.2cm}
%  \caption{Performance of learned policies $\pi_\theta$}
% \end{table}

\paragraph{\textbf{Out-of-sample behavior}} In this experiment, we evaluate the performance of the threshold-based selection criterion, defined as $j(x) = 1[x > \bar{x}]$, and compare it against standard \texttt{IPS} and \texttt{LS} baselines. We consider three thresholds $\bar{x}_1, \bar{x}_2, \bar{x}_3$, corresponding to relative improvements $I = H_n(\pi_\theta) / H_n(\pi_0) - 1$ exceeding $10\%$, $20\%$, and $30\%$, respectively. In our setup, the logging policy $\pi_0$ has an expected reward of approximately $\mathbb{E}_{\pi_0}[r] \approx 0.05$. Optimizing each criterion amounts to finding a policy $\pi_\theta$ that maximizes the probability of achieving the desired level of improvement. We simulate $100$ independent A/B tests. In each run, we collect $N = 1000$ samples, learn the various policies, and evaluate their true expected rewards. This allows us to compute the mean and the median performance, and also the probability of surpassing the specified improvement thresholds for each method. The results are reported in Table~\ref{tab:results}.

We find that directly optimizing the probability of improvement yields robust policies: they achieve high average rewards, better median outcomes, and consistently satisfy the improvement criteria. The \texttt{LS} method is competitive in terms of average performance, but slightly underperforms on the median and probability-based metrics. In contrast, the naive \texttt{IPS} baseline proves unreliable—it underperforms across all metrics and frequently selects policies that perform worse than the logging policy $\pi_0$, making it unsuitable for high-stakes decision-making scenarios.

\section{Conclusion}

This preliminary work introduces a new policy optimization method. 
We pinpoint that, like in~\cite{betlei2024maximizing}, the Algorithm~\ref{alg:COGC} can be extended to outcomes that are multidimensional, which allows to account to, for instance, budget constraints. 
The encouraging empirical results call for a more foundational understanding of the approach, which is why 
we plan to investigate  conditions under which the algorithm possesses theoretical guarantees.

% mathematical analysis is missing
% strong algorithm 
%\lipsum[1]

\newpage
\printbibliography

\end{document}